\documentclass[conference]{IEEEtran}
\IEEEoverridecommandlockouts
\usepackage{cite}
\usepackage{amsmath,amssymb,amsfonts}
\usepackage{algorithmic}
\usepackage{graphicx}
\usepackage{textcomp}
\usepackage{xcolor}

\usepackage[T1]{fontenc}
\usepackage[utf8]{inputenc}
\usepackage{hyperref}  % hyperref is not available in IEICE Transactions
\usepackage{url}
\usepackage{booktabs}
\usepackage{wrapfig}

\definecolor{navy}{RGB}{0,0,127}
\hypersetup{
  colorlinks=true,
  linkcolor=navy,
  citecolor=navy,
  filecolor=navy,
  urlcolor=navy,
linktocpage}
\newcommand{\algoref}[1]{\hyperref[#1]{Algorithm.~{\ref*{#1}}}}

\begin{document}

\title{
A Simple, Solid, and Reproducible Baseline for Bridge Bidding AI
\thanks{
\tiny{
$^*$Equal contributions.
979-8-3503-5067-8/24/\$31.00~\copyright2024 IEEE.
Personal use of this material is permitted.  Permission from IEEE must be obtained for all other uses, in any current or future media, including reprinting/republishing this material for advertising or promotional purposes, creating new collective works, for resale or redistribution to servers or lists, or reuse of any copyrighted component of this work in other works.
This study was supported by NEDO~(JPNP20006) and by JSPS KAKENHI~(No. 22H04998 and 23H04676), Japan.
}}}

\author{
\IEEEauthorblockN{Haruka Kita$^*$}
\IEEEauthorblockA{
    \textit{Kyoto University} \\
    Kyoto, Japan\\
    hrkkt1213@gmail.com}
\and
\IEEEauthorblockN{Sotetsu Koyamada$^*$}
\IEEEauthorblockA{\textit{ATR, Kyoto University} \\
Kyoto, Japan \\
koyamada@atr.jp}
\and
\IEEEauthorblockN{Yotaro Yamaguchi}
\IEEEauthorblockA{
\textit{LY Corporation}\\
Tokyo, Japan \\
yyamaguchi643@gmail.com}
\and
\IEEEauthorblockN{Shin Ishii}
\IEEEauthorblockA{
\textit{Kyoto University} \\
Kyoto, Japan\\
ishii@i.kyoto-u.ac.jp}
}

\maketitle

\begin{abstract}
    Contract bridge, a cooperative game characterized by imperfect information and multi-agent dynamics, poses significant challenges and serves as a critical benchmark in artificial intelligence (AI) research. Success in this domain requires agents to effectively cooperate with their partners. 
    This study demonstrates that an appropriate combination of existing methods can perform surprisingly well in bridge bidding against WBridge5, a leading benchmark in the bridge bidding system and a multiple-time World Computer-Bridge Championship winner.
    Our approach is notably simple, yet it outperforms the current state-of-the-art methodologies in this field. 
    Furthermore, we have made our code and models publicly available as open-source software. 
    This initiative provides a strong starting foundation for future bridge AI research,
    facilitating the development and verification of new strategies and advancements in the field.
\end{abstract}


\section{Introduction}
\label{sec:intro}
Throughout the history of artificial intelligence (AI) research, games have played pivotal roles as benchmarks for measuring progress. 
AIs have now achieved or even surpassed the skill levels of human experts in a variety of classic games.
Notable examples include backgammon~\cite{Tesauro1995}, chess~\cite{Silver2018}, Go~\cite{Silver2016,Silver2017,Silver2018}, poker~\cite{Morav_vc_ik2017,Brown2018,Brown2019}, mahjong~\cite{Li2020}, and Atari~2600~\cite{Mnih2015}.

Contract bridge joins the ranks of these classic games as a significant benchmark for AI~\cite{Ginsberg1999,Ventos2017,Yeh2018,Rong2019,Tian2020,Lockhart2020}. 
It presents complex sets of challenges due to its multi-agent nature, the imperfect information available to players, 
and the need for both cooperation within teams and competition against the opposing team. 
Bridge is somewhat akin to the game of Hanabi~\cite{Bard2020}, where information sharing is crucial, though bridge also incorporates the competitive element of playing against another team like DouDiZhu~\cite{Zha2021}.
Despite extensive research efforts, to our best knowledge, no AI has yet been demonstrated to consistently outperform top human players in bridge.

The game of bridge is structured around two main phases: bidding and playing. 
The bidding phase, in particular, is critical to success in the game~\cite{Yeh2018} and is the focus of our study. 
Our contributions to this area are twofold:
\begin{itemize}
    \item We have discovered that a straightforward integration of existing techniques can achieve state-of-the-art~(SOTA) performance in the bidding phase, specifically in tests against WBridge5\footnote{\url{http://www.wbridge5.com/}}. This program is a multiple-time winner of the World Computer-Bridge Championship (2005, 2007, 2008, and 2016-2018) and serves as the standard benchmark for bridge AI research.
    \item To foster further advancements in the field, we have made our code and trained models open-source. This allows our work to be easily reproduced and verified by others, offering a new baseline for future research in bridge AI, beyond the traditional evaluations using WBridge5.
\end{itemize}

\section{Background: Contract Bridge Overview}
Here, we provide a simplified overview of the game's flow rather than detailing all its rules.
Bridge is a card game for four players, divided into two teams. Each player receives 13 cards from a standard 52-card deck, and these cards are kept secret from the other players. The game unfolds in two main stages: the bidding phase and the playing phase.
\begin{itemize}
\item \textbf{Bidding phase.}
In this auction-style stage, players predict how many tricks (sets of four cards, one from each player) their team can win, using bids as a form of communication to signal their hand's strength and potential to their partner.
Additionally, they select a suit to serve as trump, which can override other suits to win tricks.
They make bids to set a ``contract,'' which outlines the number of tricks the team aims to win and identifies the ``declarer''~(the player who made the bid that established the final contract).
\item \textbf{Playing phase.}
Players take turns playing one card at a time, with the highest card of the led suit or trump winning the trick.
This process repeats for all 13 tricks.
\end{itemize}
The team's score depends on meeting or exceeding their contract in tricks won, with penalties for falling short.
Effective communication and strategy are key, as players must signal their hand's potential to their partner through their bids to form a winning contract.

\section{Related Work}
While advancements like those by Jack\footnote{\url{https://www.jackbridge.com/eindex.htm}}, WBridge5, and in the work of Ginsberg et al.\cite{Ginsberg1999} have seen AI reach human-level performance in the \emph{playing} phase, the \emph{bidding} phase remains a more formidable challenge\cite{Yeh2018}. This complexity has guided much of the recent focus towards improving AI performance in the bidding aspect of bridge:
Yeh et al.~\cite{Yeh2018} pioneered the application of neural networks to bridge bidding, albeit under some simplified conditions such as a restricted number of bids and opponents that always pass.
Rong et al.~\cite{Rong2019} developed a neural network-based bidding system free from these constraints. Their approach included both a policy network for decision-making and an estimation network to predict unseen hands, initially trained on data from human experts and later refined through reinforcement learning (RL) and self-play.
Gong et al.\cite{Gong2019} were the first to claim the creation of a strong bidding system developed without relying on human game data, achieving significant improvements over WBridge5. They utilized the A3C algorithm\cite{Mnih2016} for training their policy-value network entirely through self-play.
Tian et al.~\cite{Tian2020} introduced a joint policy search (JPS) algorithm tailored for cooperative games, offering theoretical assurances that JPS-derived policies would at least match the performance of baseline strategies in purely cooperative settings. Despite these guarantees not strictly applying to bridge, their application of JPS led to enhanced bidding strategies.
Lockhart et al.~\cite{Lockhart2020} focused on developing AI policies capable of cooperating with human players, achieving SOTA results against WBridge5 through the use of search techniques and policy iteration on a pretrained model. To the best of our knowledge, their work represents the current benchmark in AI performance for bridge bidding.
These studies collectively underscore the evolving landscape of AI research in bridge, highlighting a shift from foundational models to sophisticated strategies capable of navigating the game's intricate dynamics.

\section{Methods: Training Recipe}
\label{sec:methods}
This section outlines the training process for our bridge bidding model, which involves two main stages:
\begin{itemize}
\item Initially, we pretrain the neural network using supervised learning (\textbf{SL}). 
Further information is given in \autoref{sec:sl}.
\item Next, we enhance the model using the Proximal Policy Optimization (PPO) algorithm~\cite{Schulman2017}, a popular reinforcement learning (\textbf{RL}) method, combined with fictitious self-play (\textbf{FSP})~\cite{Heinrich2015}. 
Details are provided in \autoref{sec:rl}.
\end{itemize}

\subsection{Network Architecture and Input Features}
Our model processes a 480-dimensional binary input vector, 
consistent with standards set by OpenSpiel~\cite{Lanctot2019} and Pgx~\cite{Koyamada2023}. 
The input features are detailed in \autoref{tab:input}. 
The network architecture comprises a 4-layer multi-layer perceptron (MLP), 
each layer containing 1024 neurons and employing ReLU activation functions~\cite{Glorot2011}, 
following the design of Lockhart et al.~\cite{Lockhart2020}.
Outputs include a policy head for 38 actions (35 bids, pass, double, redouble) and a value head.
\begin{table}[t]
\centering
\caption{Input features.}
\label{tab:input}
\begin{tabular}{lr}
\toprule
Feature & Size \\
\midrule
Vulnerability & 4 \\
Pass before the opening bid & 4 \\
For each bid, who made it? \hfill (35 4-dim one-hot vector) & 140 \\
For each double, who made it? \hfill (35 4-dim one-hot vector) & 140 \\
For each redouble, who made it? \hfill (35 4-dim one-hot vector) & 140 \\
Current player's hand & 52 \\
\midrule
Total & 480 \\
\bottomrule
\end{tabular}
\end{table}

\subsection{Model Pretraining by Supervised Learning (SL)}
\label{sec:sl}
Initial training utilizes a dataset from OpenSpiel\footnote{\url{https://console.cloud.google.com/storage/browser/openspiel-data/bridge}}, also employed by Lockhart et al.~\cite{Lockhart2020}. 
This dataset, generated with WBridge5 but based on the SAYC bidding system\footnote{\url{https://web2.acbl.org/documentlibrary/play/SP3\%20(bk)\%20single\%20pages.pdf}}, a simple bidding system different from WBridge5's own system. 
It includes 1M boards for training and 10K for evaluation, with 12.8M state-action pairs for training and 110K for evaluation. 
We used Adam~\cite{Kingma2015} with a learning rate of $1.0 \times 10^{-4}$ and a batch size of 128, running the training over 40 epochs.

\subsection{Reinforcement Learning (RL)}
\label{sec:rl}
For model enhancement, we applied the PPO algorithm~\cite{Schulman2017}, effective in cooperative multi-agent settings~\cite{Yu2022}, and includes A2C as a special case~\cite{Huang2022a}. 
To mitigate policy cycling common in self-play, we incorporated FSP~\cite{Heinrich2015}, 
which samples the opponent uniformly from the checkpoints.

\textbf{Reward function.}
Non-zero rewards are assigned only at the end of each game.
The reward $z$ is calculated by $z = \text{score} / 7600$, 
where the score is derived from the double dummy solver~(DDS)\footnote{\url{https://github.com/dds-bridge/dds}}, 
a standard approximator for the playing phase,
and 7600 represents the maximum absolute score.

\textbf{DDS dataset.}
To bypass real-time DDS calculations during RL, we used a precomputed DDS dataset from Pgx~\cite{Koyamada2023}, 
containing 12.5M boards for training and 100K for evaluation.

\textbf{Invalid action masking.}
This technique, aimed at preventing the agent from selecting illegal actions, has been widely adopted in AI research; including notable implementations like Suphx~\cite{Li2020}, OpenAI Five~\cite{Berner2019}, and AlphaStar~\cite{Vinyals2019}, among others. For detailed insights, see~\cite{Huang2022b}.

\textbf{Other details.}
Our PPO implementation is a fork of PureJaxRL\footnote{\url{https://github.com/luchris429/purejaxrl}}~\cite{Lu2022}.
After conducting preliminary tests without using the test DDS data,
we established the following hyperparameters: 
8192 vectorized environments, 
a rollout length of 32, 
GAE $\lambda$ of 0.95, 
a discount factor of 1.0, 
a clip ratio of 0.2, 
a value loss coefficient of 0.5, 
an entropy coefficient of $1.0 \times 10^{-3}$, 
a batch size of 1024, 
using Adam, 
with a learning rate of $1.0 \times 10^{-6}$. 
We trained the model for $10^4$ PPO update steps, in which each step has 10 epochs over rollout data.
% The model went through 10 epochs over the DDS dataset.

\section{Results}

\subsection{Performance against WBridge5}

To assess our model's effectiveness, trained as described in \autoref{sec:methods}, 
we tested it against WBridge5, the leading benchmark in computer bridge. 
We utilized WBridge5 at its highest difficulty setting and with its native bidding system, 
which differs from the SAYC system used during our SL pretraining phase. 
The evaluation comprised 1K games, conducted over a day, 
reflecting the significant time needed because WBridge5 operates with a GUI and includes a playing phase.

The outcomes, detailed in \autoref{tab:results}, 
also compare our model's performance with that reported in prior studies. 
Our approach achieved an average of +1.24 International Match Points (IMPs)\footnote{Established in Law 78B: \url{https://web2.acbl.org/documentlibrary/play/laws-of-duplicate-bridge.pdf}.} per board against WBridge5 across these games, 
surpassing the previous SOTA performance of +0.85 IMPs/board by Lockhart et al.~\cite{Lockhart2020}. 
This improvement of 0.39 IMPs/board is significant in the context of computer bridge competitiveness~\cite{Ventos2017}.
\begin{table}[t]
    \caption{Performance against WBridge5.}
    \label{tab:results}
    \centering
    \begin{tabular}{llr}
        \toprule
        Paper & IMPs/board ($\pm$SE) & \# games\\
        \midrule
        Rong et al. \cite{Rong2019} & $+0.25$ ($\pm$N/A) & 64 \\
        Gong et al. \cite{Gong2019} & $+0.41$ $(\pm 0.27)$ & 64 \\
        Tian et al. \cite{Tian2020} & $+0.63$ $(\pm 0.22)$ & 1K \\
        Lockhert et al. \cite{Lockhart2020} & $+0.85$ $(\pm 0.05)$ & 10K \\
        \midrule
        Ours & $\mathbf{+1.24}$ $(\pm 0.19)$ & 1K \\
        \bottomrule
    \end{tabular}
\end{table}

\subsection{Ablation Study}
\begin{figure}[t]
    \centering
    \includegraphics[width=0.85\linewidth]{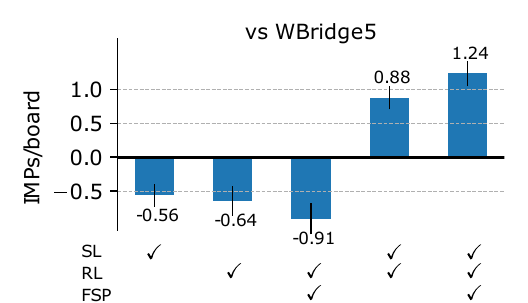}
    \caption{Ablation of each training component.}
    \label{fig:ablation}
\end{figure}

Our method combines \textbf{SL} pretraining with \textbf{RL} model improvement through \textbf{FSP}. 
To dissect the contribution of each component, 
we tested variations of our model lacking one of these elements against WBridge5, 
with findings summarized in \autoref{fig:ablation}.
We used a learning rate 10 times larger for the model from scratch~(i.e., w/o SL),
as we found that it performs better than the original learning rate in those settings.
We also trained the model from scratch with twice the number of steps to compensate for the lack of SL pretraining.

Key observations include:
\begin{enumerate}
    \item Removing SL pretraining drastically reduces performance, rendering the model unable to surpass the WBridge5 baseline.
    \item Integrating FSP enhances results post-SL pretraining but is ineffective on its own.
\end{enumerate}
The first insight challenges Gong et al.'s~\cite{Gong2019} assertion that a model can outperform WBridge5 without SL pretraining, 
a claim we could not replicate despite extensive hyperparameter testing. 
We leave further exploration of this discrepancy for future work.
We can offer a plausible explanation for the second observation. 
Starting from scratch, facing a random~(or nearly random) opponent policy might slow the learning process.
It is important to note that the bidding system used to create the dataset for SL pretraining differs from WBridge5's system. 
Therefore, the model enhanced with FSP is not just learning to outperform a version that mimics WBridge5.

\begin{figure}[t]
    \centering
    \includegraphics[width=0.92\linewidth]{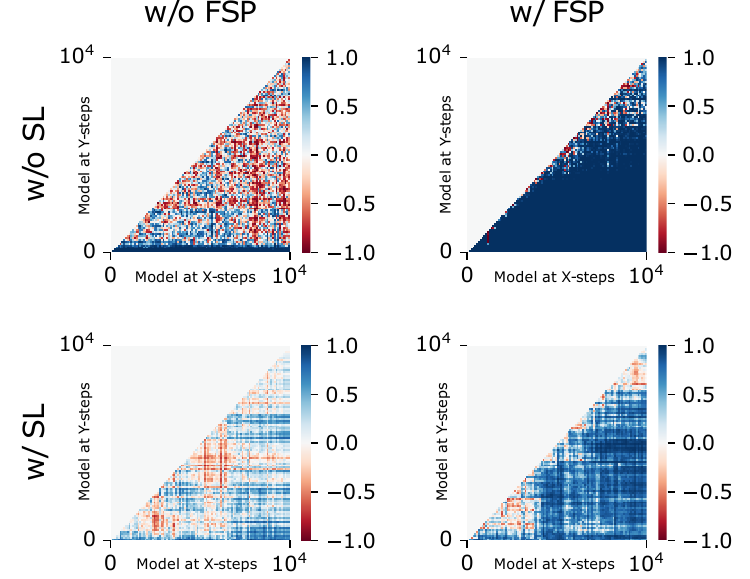}
    \caption{
        Comparison of usual self-play (w/o FSP) and FSP.
        Each item represents the IMPs/board scaled by tanh of the model at the X-steps against the model at the Y-steps~(X is greater than Y).
        }
    \label{fig:sp-vs-fsp}
\end{figure}

To verify the mitigation of policy cycling by FSP, 
we organized a round-robin tournament among training checkpoints.
\autoref{fig:sp-vs-fsp} shows the results.
Unlike standard self-play, where some later-stage models might struggle against earlier ones, 
FSP consistently demonstrated the ability to outperform its predecessors, 
underscoring its value in stable training.

\section{Open-Source Software and Models}
Our straightforward approach, as detailed in \autoref{sec:methods}, has demonstrated SOTA performance against the most recognized benchmark in computer bridge. While effective, this method is not specifically optimized for bridge's unique aspects, indicating potential areas for enhancement. To encourage continued advancement in bridge AI research, we are releasing our code and trained models as open-source resources:

\begin{center}
\url{https://github.com/harukaki/brl}
\end{center}

This new baseline aims to overcome certain limitations associated with the current WBridge5 benchmark:
\begin{enumerate}
\item \textbf{Slow WBridge5 evaluation.} Primarily designed for human interaction, WBridge5's evaluation process, which relies on GUI operations and includes a playing phase, is notably time-consuming and resource-intensive. This was highlighted by Rong et al.~\cite{Rong2019}, who manually tested their model against WBridge5.
\item \textbf{Potential weakness of WBridge5.} As evidenced in \autoref{tab:results}, recent advancements have significantly outperformed WBridge5, raising questions about the benchmark's current competitiveness. Moreover, fairness in evaluation is a concern since WBridge5 does not incorporate DDS strategies, although recent studies trained their models with DDS datasets.
\end{enumerate}
By addressing these issues, our baseline not only offers a more efficient and equitable framework for assessment but also enhances the diversity of bidding systems under consideration.

\section{Limitations, Future Work, and Conclusion}
Our study demonstrates that straightforward integration of existing techniques can outperform WBridge5, a leading benchmark in computer bridge bidding systems. However, our approach relies on SL pretraining to surpass WBridge5, contrasting with Gong et al.~\cite{Gong2019}, who claimed to achieve superior results without SL, using only RL from scratch. 
Exploring the reasons behind this discrepancy presents a valuable opportunity for future research.

Additionally, our methodology, while effective, is not specifically designed with the unique aspects of bridge in mind. This suggests there may be room for further optimization and refinement tailored to bridge's strategic complexities.

Despite these limitations, we are confident our work lays a solid foundation for subsequent studies in bridge AI. 
By providing our code and models as open-source resources, we aim to facilitate the development of more advanced AI systems capable of exceeding human expertise in bridge. 

\bibliographystyle{IEEEtran}
\bibliography{main}

\end{document}